\documentclass[graybox]{svmult}

\usepackage{mathptmx}       % selects Times Roman as basic font
\usepackage{helvet}         % selects Helvetica as sans-serif font
\usepackage{courier}        % selects Courier as typewriter font
\usepackage{type1cm}        % activate if the above 3 fonts are
\usepackage{makeidx}         % allows index generation
\usepackage{graphicx}        % standard LaTeX graphics tool
\usepackage{multicol}        % used for the two-column index
\usepackage[bottom]{footmisc}% places footnotes at page bottom

\usepackage[numbers]{natbib}

\usepackage{tikz}
\definecolor{kit-green100}{rgb}{0,.59,.51}
\definecolor{kit-green70}{rgb}{.3,.71,.65}
\definecolor{kit-green50}{rgb}{.50,.79,.75}
\definecolor{kit-green30}{rgb}{.69,.87,.85}
\definecolor{kit-green15}{rgb}{.85,.93,.93}
\tikzstyle{mybox} = [draw=black, very thick, rectangle, rounded corners, inner sep=1pt, inner ysep=5pt]
\tikzstyle{fancytitle} = [fill=kit-green100, text=white]
\tikzstyle{myArrow} = [->, thick]

\usepackage{amsmath}
\usepackage{amsfonts}
\DeclareFontEncoding{LS1}{}{}
\DeclareFontSubstitution{LS1}{stix}{m}{n}
\DeclareSymbolFont{arrows1}{LS1}{stixsf}{m}{n}
\DeclareMathDelimiter{\barleftarrowrightarrowbar}{\mathord}{arrows1}{"C0}{arrows1}{"C0}

\usepackage{tcolorbox}

\usepackage{threeparttable}

\newcommand{\MyBilingualism}[2]{#2}

\makeindex             % used for the subject index
\begin{document}
\newcommand{\KIT}{Karlsruhe Institute of Technology, Institute for Anthropomatics and Robotics, Karlsruhe, Germany}
\newcommand{\FirstAuthor}{Stefan Constantin}
\newcommand{\SecondAuthor}{Jan Niehues}
\newcommand{\ThirdAuthor}{Alex Waibel}
\newcommand{\TitleName}{An End-to-End Goal-Oriented Dialog System with a Generative Natural Language Response Generation}
\newcommand{\AbstractText}{Recently advancements in deep learning allowed the development of end-to-end trained goal-oriented dialog systems.
Although these systems already achieve good performance, some simplifications limit their usage in real-life scenarios.

In this work, we address two of these limitations: ignoring positional information and a fixed number of possible response candidates.
We propose to use positional encodings in the input to model the word order of the user utterances.
Furthermore, by using a feedforward neural network, we are able to generate the output word by word and are no longer restricted to a fixed number of possible response candidates.
Using the positional encoding, we were able to achieve better accuracies in the Dialog bAbI Tasks and using the feedforward neural network for generating the response, we were able to save computation time and space consumption.
}

\title*{\TitleName}
\titlerunning{A Dialog System with a Generative NLG}
\author{\FirstAuthor \and \SecondAuthor \and \ThirdAuthor}
\institute{\FirstAuthor \and \SecondAuthor\and \ThirdAuthor
\at \KIT \\
\email{\texttt{firstname.lastname@kit.edu}}}
\maketitle

\abstract{\AbstractText}
% \abstract* does not appear in the paper it is only taken for SpringerLink.com, abstract does appear in the paper; Have I use abstract?
\abstract*{\AbstracText}

% \chapter*{\TitleName}

\section{Introduction}
\label{sec:Introduction}
A goal-oriented dialog system should fulfill the user's request.
If there are ambiguities or missing information in the request, the system should ask for a clarification or for the missing information.
In such a dialog system, there can be a priori knowledge about the domain.
However, dialog systems without any a priori knowledge about the domain and which are only trained with a training dataset make for a very interesting research topic.
They allow extending or changing the domain only by training them with an extended or a changed training dataset, respectively.
Recently advancements in deep learning allow the development of such end-to-end trained goal-oriented dialog systems.
No human written rules are necessary.
\cite{BordesW16} describes an end-to-end goal-oriented dialog system without a priori knowledge based on Memory Networks, which achieved good results.

This dialog system ignores the position of the words in an utterance.
That means, the encoded utterance of the utterance ``I'm in the office and not in the production hall'' is equal to the encoded utterance of the utterance ``I'm in the production hall and not in the office''.
This work presents an encoding that encodes the position of the words of an utterance.

The dialog system described in \cite{BordesW16} generates a response by predicting the best response from a fixed set of response candidates.
This approach has some performance drawbacks.
The computation time and the space consumption increase linearly with the number of response candidates.
This work presents an approach that eliminates these drawbacks by generating a response word by word.

Furthermore, this work investigates how sensitive the dialog system to user input that differ from the fictional user input in the training dataset is.

\section{Related Work}
\label{sec:RelatedWork}
In \cite{BordesW16}, an end-to-end goal-oriented dialog system based on Memory Networks with good results is presented.
Memory Networks are described in general in \cite{WestonCB14} and the end-to-end trainable Memory Network variant is described in \cite{SukhbaatarSWF15}.

To train a dialog system without a priori knowledge for a domain, a good training dataset is needed.
In \cite{SerbanLCP15}, more than 50 datasets are presented.
Only four datasets are presented in \cite{DodgeGZBCMSW15}, but the information on which skill of a dialog system the dataset is testing is added to every dataset.
Datasets can be created by humans in a fashion in which how they would be using the system or synthetically from patterns.
Synthetic datasets often do not have so many variances than real dialogs, but they are easier to create.
In \cite{WestonBCM15}, the intention for synthetic datasets is given.
If a dialog system cannot handle synthetic datasets, it will not be able to handle real datasets.
The opposite, in general, is false.

For the Natural Language Understanding (NLU) component and the Dialog Manager (DM) component there are also other appropriate neural network architectures besides Memory Networks: the Recurrent Entity Networks \cite{HenaffWSBL16}, the Differentiable Neural Computer \cite{GravesWRHDGCGRA16}, the Neural Turing Machine \cite{GravesWD14}, and the Dynamic Memory Networks \cite{KumarISBEPOGS15}.

This work uses a feedforward neural network (FNN) for Natural Language Generation (NLG).
A promising approach for an end-to-end trainable NLG is the semantically controlled long short-term memory (SC-LSTM), which is presented in \cite{WenGMSVY15}. It is a further development of the long short-term memory (LSTM).

\section{Dialog System}
The structure of the dialog system is based on \cite{BordesW16}.
Section \ref{sec:DialogSystemBase} briefly describes the structure of the dialog system described in \cite{BordesW16}.
All the equations are from \cite{BordesW16}.
Section \ref{sec:DialogSystemImprovements} presents the improvements: position encoding and temporal encoding.
The NLG word by word approach is presented in Section \ref{sec:DialogSystemNLG}.

\subsection{Foundation}
\label{sec:DialogSystemBase}
A dialog \(D\) is split into subdialogs \(D_i\).
The first subdialog \(D_1\) includes the first utterance of the user \(c_1^u\) and the first utterance of the system \(c_1^r\).
Every subdialog \(D_i\) includes its predecessor \(D_{i-1}\) and the i-th utterance of the user and of the system.
The subdialogs are split into three parts: the last utterance of the user \(c_i^u\), the last utterance of the system \(c_i^r\) and the other utterances compose the history.

Each utterance of the history is encoded in a \(V\)-dimensional vector.
\(V\) is the size of the vocabulary.
Every element of the vector corresponds to a different word in the vocabulary.
An element has the value 1 if the corresponding word occurs in the utterance; otherwise it has the value 0.
This encoding is called bag-of-words (BOW) encoding.
The mapping from an utterance to such a \(V\)-dimensional vector is called \(\Phi\).

The vectors are multiplied by a \(d \times V\)-dimensional matrix \(A\) and saved as \(d\)-dimensional utterance-embeddings in the memory:
\[m = (A\Phi(c_1^u), A\Phi(c_1^r)..., A\Phi(c_{t-1}^u), A\Phi(c_{t-1}^r))\]
In the same way, the last utterance of the user \(c_i^u\) is calculated to an utterance-embedding \(q\).

The relevance \(p_i\) of a memory entry \(m_i\) to \(q\) is calculated in the following manner:
\[p_i = Softmax(q^T m_i)\]
Every memory entry is multiplied by its relevance and then these results are summed up.
The sum is multiplied by a matrix \(R\) and this result is the vector \(o\):
\[o = R \sum_i p_i m_i\]
The output of the Memory Network is the sum of \(q\) and \(o\).

The described Memory Networks can be stacked together.
The sum of \(q_{h}\) and \(o_{h}\) is used as the new input \(q_{h+1}\) for the next Memory Network in the stack.
The number of the Memory Networks in the stack is called \(N\) and the output of the last Memory Network is \(q_{N+1}\).

The temporal order of the utterances is directly encoded in the conversation.
Therefore, \(t\) keywords are defined, which are only allowed to be used for the temporal encoding.
The keywords are added to the vocabulary.
To every utterance, the keyword that encodes how many utterances were made before this utterance is added.
The keyword for the \(t\)-th dialog before this utterance is used for the utterances before as well.

For the NLG, a candidates' selection approach is used.
The result of the multiplication of the output of the Memory Network \(q_{N+1}\) by a trainable matrix \(W\) is multiplied by all the \(C\) encoded response candidates \(y_i\):
\[\hat{a} = Softmax({q_{N+1}}^T W \Phi(y_1),..., {q_{N+1}}^T W \Phi(y_C))\]
The result \(\hat{a}\) is a \(C\)-dimensional vector.
The corresponding candidate to the element of \(\hat{a}\) that has the highest value is the predicted response.

\subsection{Position and Temporal Encoding}
\label{sec:DialogSystemImprovements}
To retain the order of the words in the utterances, a position encoding is described in \cite{SukhbaatarSWF15}.
A memory entry \(m_i\) is calculated with this encoding in the following manner:
\[m_i = \sum_j l_j \cdot Ax_{ij},\; l_{kj} = (1 -j/J)-(k/d)(1-2j/J)\]
\(J\) is the maximum utterances length, \(j\) the index of the regarded word, \(k\) the index of the regarded element of the embedding vector, and \(x_{ij}\) is the j-th word of the i-th utterance.
\newline
\newline
\cite{SukhbaatarSWF15} presents an alternative to the temporal encoding with keywords.
A trainable matrix \(T_A\) is used to calculate the utterance-embeddings of the memory entries.
The calculation of \(m_i\) with the position encoding and the temporal encoding is:
\[m_i = \sum_j l_{j} \cdot Ax_{ij} + T_A(i)\]
\(T_A(i)\) is the i-th row of \(T_A\).
In this paragraph, the number of rows of \(T_A\) is called \(t\).
With the temporal encoding only dialogs with a maximum length of \(t\) are supported.
This alternative approach is not used for the dialog system in this work, because the dialog system in this work should not have this constraint.
A workaround for this constraint is to use \(T_A\) for the last \(t\) utterances of the history and \(T_A(0)\) for all the utterances before or to remove old utterances from the history.

\subsection{NLG word by word approach}
\label{sec:DialogSystemNLG}

For every response prediction the result of the multiplication output of the Memory Network by the matrix \(W\) must be multiplied by every encoded response candidate.
Therefore, the computation time of the NLG is linear in the number of the response candidates.
Since all response candidates must be saved, the space consumption is also linear in the number of response candidates.
To eliminate these drawbacks, a word by word approach is developed for this work.
This approach uses an FNN.

The input for the FNN in timestep \(t\) is the output of the Memory Network \(q_{N+1}\) and the word-embeddings of the last \(m\) words \(w_t\), \(w_{t-1}\), ..., \(w_{t-m+1}\). 
For words with an index smaller than 1, the word-embedding of a keyword that must only be employed for this case is used.
The variable \(m\) is a hyperparamter.
The size of \(m\) is especially relevant in the utterances in which words occur multiple times.
In the case where \(m\) has the size 1, there is, for example, a problem in the utterance ``Do you prefer a water or a cola?''.
The FNN after the output of one of the two ``a'' in both cases has the same input.
The output of the Memory Network is constant and the last outputted word is ``a''.
This leads to a result in which at least one output of the FNN is false.
Therefore, \(m\) should at least have the size of the longest repeating n-gram that can occur in a possible response.

\begin{figure}
\centering

\begin{tikzpicture}[x=.5cm, y=.5cm,domain=0:9,smooth]
\node [mybox] at (0, 0.15\textheight) (uk){
    \begin{minipage}[c][0.125\textheight]{0.1\textwidth}
        \center \(q_{h+1}\)
    \end{minipage}
};
\node [mybox] at (0, 0) (wt){%
    \begin{minipage}[c][0.125\textheight]{0.1\textwidth}
        \center \(w'_t\)
    \end{minipage}
};
\node [mybox] at (0.3\textwidth, 0.0625\textheight) (h1){%
    \begin{minipage}[c][0.25\textheight]{0.1\textwidth}
        \center \(h_t\)
    \end{minipage}
};

\node [mybox] at (0.7\textwidth, 0.0625\textheight) (wt1){%
    \begin{minipage}[c][0.35\textheight]{0.1\textwidth}
        \center \(w_{t+1}\)
    \end{minipage}
};

\draw [myArrow] (uk) -- (h1) node [midway, above, sloped] (TextNode) {\MyBilingualism{voll vernetzt}{fully connected}};
\draw [myArrow] (wt) -- (h1) node [midway, below, sloped] (TextNode) {\MyBilingualism{voll vernetzt}{fully connected}};
\draw [myArrow] (h1) -- (wt1) node [midway, above, sloped] (TextNode) {\MyBilingualism{voll vernetzt mit Bias}{fully connected with bias}};

\draw [myArrow] (wt1.south) to[bend left] node [midway, below] {\MyBilingualism{Wort-Embedding von \(w_{t+1}\) ergibt \(w'_{t+1}\)}{word-embedding of \(w_{t+1}\) results in \(w'_{t+1}\)}} (wt.south);
\end{tikzpicture}

\caption{NLG word by word approach with an FNN (\(m = 1\))}
\label{fig:FNN}
\end{figure}
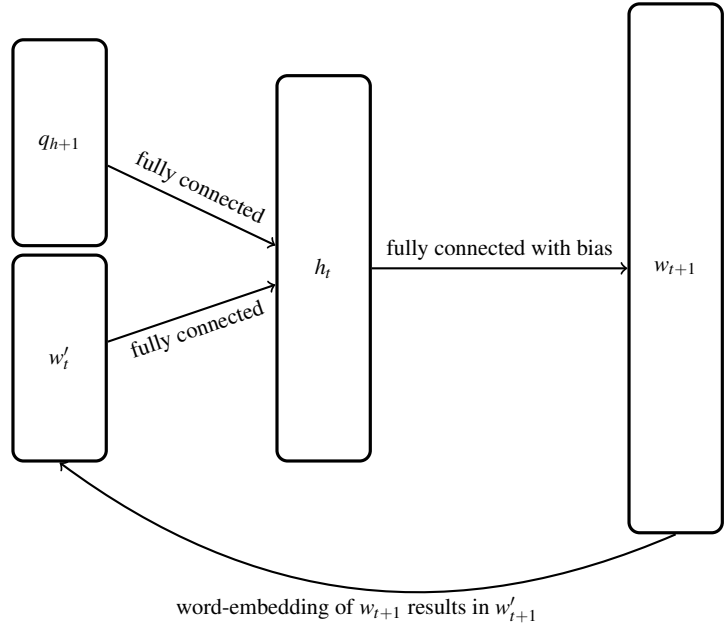

The input of the FNN is fully connected with a hidden layer.
The number of neurons is variable.
The output layer has as many neurons as there are different words in the vocabulary.
Every word is clearly allocated to a neuron.
The word that belongs to the neuron with the highest value is the word that is outputted.
The output layer is fully connected with the hidden layer and every neuron of the output layer has a bias.
The softmax function is applied to the output values.
To the outputted word \(w_{t+1}\), a word-embedding \(w'_{t+1}\) is calculated.
This word-embedding is used in the next time step \(t+1\) as the input for the FNN.
The described structure of the FNN is depicted in Figure \ref{fig:FNN}.

The weights of the FNN are randomly and equally initialized in the interval -1 to 1.
Without this large span, the FNN delivers bad results.
The FNN is used as many times as the many words that the longest outputted utterance should have.
In the case that a response is shorter, after the last word of this response, a keyword that must be only used in this case is outputted.
The FNN builds together with the Memory Network, which calculates \(q_{N+1}\), a neural network.
The whole neural network can be trained end-to-end.
The weights of the Memory Network are normally distributed with an expected value of 1 and a standard derivation of 0.1.
The error function is cross entropy and the Adam optimizer is used.
The batch size is 32.

\section{Results}
\label{sec:Results}
The dialog system described in \cite{BordesW16} was evaluated with the Dialog bAbI Tasks.
These tasks are also described in \cite{BordesW16} and contain six tasks.
Task 1 tests if a dialog system can issue an API call with four parameters.
The return of the API call is the return of a database query.
The user gives 0, 1, 2, 3 or 4 parameters and the dialog system must ask for the missing parameters.
Task 2 tests if after issuing the API call a new API call with updated parameters can be issued.
The ability to select the best element from the returned elements of the database query is tested by task 3 and the ability to give information about a certain element of the return of the database query is tested by task 4.
Task 5 combines the first four tasks.
Real dialogs from the the Second Dialog State Tracking Challenge \cite{HendersonTW2014} was formatted in the format of the Dialog bAbI Tasks and constitute task 6.

To compare the performance of the dialog system described in this work with the dialog system described in \cite{BordesW16}, the Dialog bAbI Tasks are used as the evaluation dataset.

The accuracies of the presented dialog system in the Dialog bAbI Tasks are depicted in Figure \ref{fig:Results}.
All accuracies are the accuracies of the test dataset, evaluated at the state after the epoch of the training with the highest accuracy in the validation dataset regarded the subdialogs.
All the tasks were trained six times with 100 epochs and there was an evaluation after every fifth epoch for the NLG candidates' selection approach.
For the NLG word by word approach there was an evaluation after every epoch.

The accuracies given in parenthesis are the accuracies for the complete dialogs, and the other value is the accuracy for the subdialogs.
\begin{figure}
\centering

\begin{threeparttable}
\begin{tabular}{p{0.9cm} | p{2cm} | p{2cm} | p{2cm} | p{2cm} | p{2cm}}
\hline
 & candidates, evaluated in \cite{BordesW16} & candidates with position encoding & word by word with position encoding & candidates with BOW encoding & word by word with BOW encoding  \\ \hline
task 1 & 99.6 (99.6) & 99.95 (99.70) & 99.98 (99.90) & 99.24 (95.70) & 100.00 (100.00) \\
task 2 & 100 (100) & 99.92 (99.20) & 99.23 (92.70) & 98.55 (86.30) & 100.00 (100.00) \\
task 3 & 74.9 (2) & 74.98 (0.00) & 74.90 (0.00) & 74.59 (0.00) & 74.90 (0.00) \\
task 4 & 59.5 (3) & 57.26 (0.10) & 57.18 (0.00) & 57.20 (0.00) & 57.18 (0.00) \\
task 5 & 96.1 (49.4) & 95.71 (46.90) & 89.07 (10.50) & 92.31 (20.50) & 86.84 (6.20) \\
task 6 & 41.1 (0) & 40.72 (0.81) & 22.26 (0.00) & 37.54 (0.18) & 21.65 (0.00) \\
task 1 * & - (-) & 99.78 (98.70) & 100.00 (100.00) & 99.19 (95.20) & 99.98 (99.90) \\ \hline
\end{tabular}

\begin{tablenotes}[para,flushleft]
* with 30\,788 dummy candidates
\end{tablenotes}
\end{threeparttable}

\caption{Accuracies in percent in the Dialog bAbI Tasks}
\label{fig:Results}
\end{figure}
For the dialog system with the NLG candidates' selection approach the following hyperparameter was chosen for the evaluation: 0.0058 for the learning rate of the Adam optimizer, 44 for the embedding-size, and 1 for the number of stacked Memory Networks.
The NLG word by word approach has the following hyperparameter: 0.0022 for the learning rate of the Adam optimizer, 59 for the embedding-size, 3 for the number of stacked Memory Networks, and 50 for the number of neurons of the hidden layer of the FNN.
The FNN got only the last outputted word as the input.
The batches were shuffled before every epoch.
\newline
\newline
It is important that a dialog system can generalize the trained dialogs.
The inputs of a real user can very likely differ from the inputs of the fictional user in the training dataset.
Here, it is investigated whether the ability to generalize can be evaluated with the Dialog bAbI Tasks.
In task 1, the training dataset does not include any of the API calls of the test dataset.
This fact and the good results of the dialog system in task 1 lead to the fact that the dialog system can build API calls autonomously.

All the utterance patterns that occur in the test dataset also occur in the training dataset.
Therefore, for this work, a dialog from task 1 of the Dialog bAbI Tasks was selected.
In this dialog the user gives 0 parameters and the dialog system asks for the missing 4 parameters.
5 user utterances were modified. 
The utterances were shortened without to change the sense of them and after that the word ``please'' was added.
The dialog, the possible parameter values and the modified utterances are depicted in Figure \ref{fig:ModifiedDialogs}.
All combinations of parameter values and one modified utterance per dialog generates 13\,500 dialogs.
In these dialogs, the NLG candidates' selection approach had an accuracy of 60.00\,\% with the BOW encoding and 60.00\,\% with the position encoding and the NLG word by word approach had an accuracy of 98.60\,\% with the BOW encoding and 94.20\,\% with the position encoding.
The accuracies are the accuracies for the complete dialogs.

\begin{figure}
original dialog:
\begin{enumerate}
\item hello\(\barleftarrowrightarrowbar\)hello what can i help you with today
\item can you book a table\(\barleftarrowrightarrowbar\)i'm on it
\item \textless SILENCE\textgreater\(\barleftarrowrightarrowbar\)any preference on a type of cuisine
\item with \emph{cuisine} food\(\barleftarrowrightarrowbar\)where should it be
\item in \emph{location}\(\barleftarrowrightarrowbar\)how many people would be in your party
\item we will be \emph{number} people\(\barleftarrowrightarrowbar\)which price range are looking for
\item in a \emph{price} price range please\(\barleftarrowrightarrowbar\)ok let me look into some options for you
\item \textless SILENCE\textgreater\(\barleftarrowrightarrowbar\)api\_call italian rome four cheap
\end{enumerate}

parameter values:\newline
cuisine: british, french, indian, italian, spanish\newline
number: two, four, six, eight\newline
location: bombay, london, madrid, paris, rome\newline
price: cheap, moderate, expensive\newline

modified user utterances:
\begin{enumerate}
\setcounter{enumi}{1}
\item book a table please\(\barleftarrowrightarrowbar\)i'm on it
\setcounter{enumi}{3}
\item \emph{cuisine} food please\(\barleftarrowrightarrowbar\)where should it be
\item \emph{location} please\(\barleftarrowrightarrowbar\)how many people would be in your party
\item \emph{number} please\(\barleftarrowrightarrowbar\)which price range are looking for
\item please \emph{price} price range\(\barleftarrowrightarrowbar\)ok let me look into some options for you
\end{enumerate}

\caption{Modified user utterances}
\label{fig:ModifiedDialogs}
\end{figure}
\begin{figure}
\centering
\includegraphics[width=\textwidth]{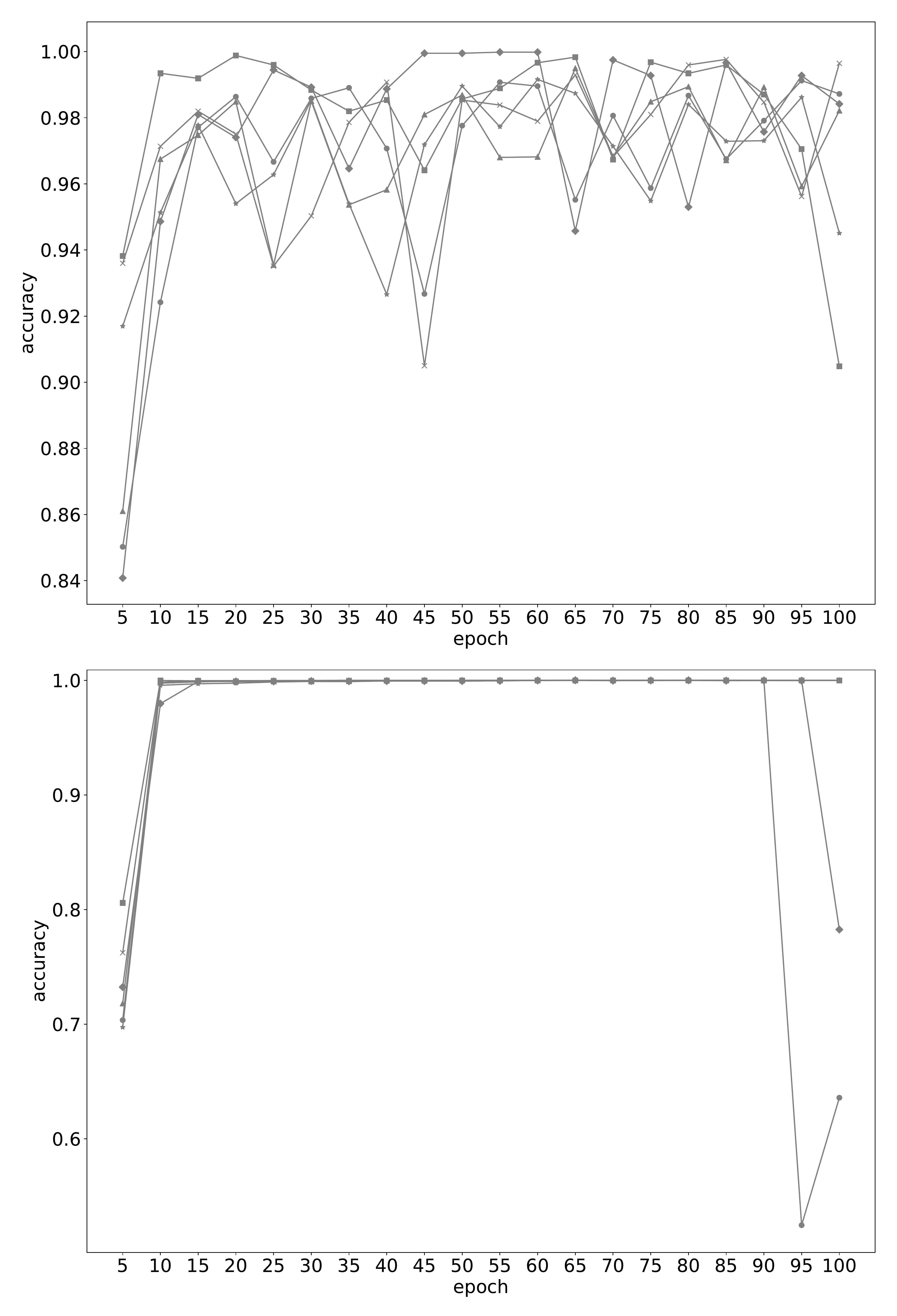}
\caption{Task 1 of the Dialog bAbI Tasks with the NLG word by word approach (above) and the candidates' selection approach (below), position encoding is used for both approaches}
\label{fig:Task1Results}
\end{figure}

~
\newline
Figure \ref{fig:Task1Results} depicts the accuracies of task 1 from the six runs of both NLG approaches with the position encoding evaluated every fifth epoch (totaling 100 epochs).
This figure should show how good the results can be reproduced.
The accuracy is the accuracy of the test dataset.
In summary, the result of the NLG candidates' selection approach has a good reproducibility.
Whereas the NLG word by word approach often needs multiple runs for good results.
These good results are often not stable but rather decrease with higher epochs.
Therefore, it is important to evaluate after every epoch to get good results and buffer the state of the neural network. 
The large span in the initialization of the weights in the FNN, as described in Section \ref{sec:DialogSystemNLG}, is probable the reason for this bad reproducibility.
\newline
\newline
In the tested datasets with a maximum of 4212 candidates the dialog system with the NLG word by word approach needed longer computation times.
By increasing the number of candidates of task 1, the dialog system with the NLG candidates' selection approach showed a worse computation time and the computation time of the dialog system with the NLG word by word approach was independent of the number of candidates.
With 4212 candidates, the NLG word by word approach needed 2.80 more computation time than the NLG candidates' selection approach.
However, with 35\,000 candidates the NLG candidates' selection approach needed 1.08 more computation time than the other NLG approach.
For all the computation times, it was used 100 epochs training of task 1, position encoding as encoding, and the training state was evaluated after every fifth epoch.
The computation time of the NLG word by word approach did not get worse by adding more candidates.
These computation times were measured by a computer with an Intel Core i7 7700 CPU, an Nvidia GeForce 1070 GTX GPU and 16 GB main memory (the swap was not used).

For the NLG candidates' selection approach, all the possible response candidates have to be multiplied by the output of the Memory Network.
The main memory consumption depends on the implementation.
More memory is needed if all the vector representations of the candidates are in the memory instead of computing the vector representation on the fly, which increases the computation time.
However, the candidates must be on the disk.
For the computation times in this work, the vector representations of all the candidates were held in the memory.

\section{Conclusions and Further Work}
\label{sec:Conlusion}
The position encoding had better accuracies than the bag-of-words encoding in the Dialog bAbI Tasks for the NLG candidates' selection approach.
However, for the NLG word by word approach with the position encoding, the accuracies were in some tasks worse.
For the experiment with utterances that was modified from the user utterances in the training dataset, the bag-of-words encoding produced a better accuracy for the NLG word by word approach and both encodings produced the equal accuracy for the NLG candidates' selection approach.

The presented NLG word by word approach showed in task 1 of the Dialog bAbI Tasks a better accuracy than the NLG candidates' selection approach.
Furthermore, the accuracy of the NLG word by word approach was higher in the experiment with the modified utterances.
The space consumption was lower with the NLG word by word approach and the computation time was lower with large numbers of candidates (more than 35\,000 response candidates).
\newline
\newline
In future work, there must be deeper research to find whether more sophisticated neural network architectures can improve the performance of the word by word approach.

The utterance-embeddings have the problem that small differences can lead to other utterance-embeddings.
This means that spelling errors of words can lead to other utterance-embeddings. 
To avoid learning all the possible spelling errors, a spelling corrector component before the Memory Network could mitigate this problem.
A normalization component can also help in reducing the necessary size of the datasets.
Future works might investigate if these proposed upstream components can lead to a better performance or if these components themselves introduce errors and lead to a lower accuracy.

%\section*{Appendix}
%\label{sec:Appendix}

\section*{Acknowledgement}
This work has been conducted in the SecondHands project which has received funding from the European Union’s Horizon 2020 Research and Innovation programme (call:H2020- ICT-2014-1, RIA) under grant agreement No 643950. This work was supported by the Carl-Zeiss-Stiftung.

\bibliographystyle{spmpsci.bst}
\bibliography{bibliography}
\end{document}